\documentclass{article}
\usepackage{spconf,amsmath,epsfig}
\usepackage[ruled,vlined]{algorithm2e}
\usepackage{fancyhdr}
\usepackage{textpos}

\fancypagestyle{firstpage}{
    \fancyhf{}
    \fancyfoot[L]{IEEE} % Left side footer
    \fancyfoot[C]{\thepage} % Centered page number
    \fancyfoot[R]{IGARSS 2024} % Right side footer
}

\fancypagestyle{otherpages}{
    \fancyhf{}
    \fancyfoot[C]{\thepage} % Centered page number
}

\newcommand{\mypm}{\mathbin{\smash{%
\raisebox{0.35ex}{%
            $\underset{\raisebox{0.5ex}{$\smash -$}}{\smash+}$%
            }%
        }%
    }%
}

\title{Assessing Annotation Accuracy in Ice Sheets Using Quantitative Metrics}
\name{Bayu Adhi Tama, Vandana Janeja\sthanks{Corresponding author}, Sanjay Purushotham}
\address{University of Maryland, Baltimore County, MD, USA}
\begin{document}
\thispagestyle{firstpage}
%\ninept
%
\maketitle
% Place vertical text on the first page
\begin{textblock*}{1cm}(-2cm,1cm) % Adjust position as needed
    \rotatebox{90}{\footnotesize This paper is accepted at the IGARSS 2024 - 2024 IEEE International Geoscience and Remote Sensing Symposium}
\end{textblock*}

\pagestyle{otherpages}
\begin{abstract}
The increasing threat of sea level rise due to climate change necessitates a deeper understanding of ice sheet structures. This study addresses the need for accurate ice sheet data interpretation by introducing a suite of quantitative metrics designed to validate ice sheet annotation techniques. Focusing on both manual and automated methods, including ARESELP and its modified version, MARESELP, we assess their accuracy against expert annotations. Our methodology incorporates several computer vision metrics, traditionally underutilized in glaciological research, to evaluate the continuity and connectivity of ice layer annotations. The results demonstrate that while manual annotations provide invaluable expert insights, automated methods, particularly MARESELP, improve layer continuity and alignment with expert labels. 
\end{abstract}
\section{Introduction}
The escalating sea level rise, propelled by the ongoing climatic changes, underscores the vital need to understand the fundamental structure of ice sheets~\cite{golledge2020long}. Acquiring this understanding is essential for enhancing the accuracy of future forecasts regarding the rise in sea levels, a topic of great concern for both coastal communities and climate researchers~\cite{dutton2015sea}. However, extracting essential information from ice sheet data is a substantial problem. The task is intricate and can be accomplished using several methodologies, such as manual, semi-automated, and fully automated procedures. Each of these solutions necessitates a substantial dedication of time and expertise from experts who meticulously annotate and assess complex data sets~\cite{delf2020comparison}.

Considering the intricate nature and importance of these interpretations, validating ice sheet annotation methods becomes a vital element of glaciological research~\cite{tack2023metrics}. Errors or omissions in annotation might result in significant inaccuracies in comprehending and predicting ice sheet dynamics. In light of the crucial requirement for precision and dependability, this study presents a collection of new quantitative measurements. These measures aim to assess the accuracy of ice sheet annotation techniques thoroughly, enhancing our comprehension of glacial dynamics and their impact on sea level fluctuations. This study aims to leverage several metrics from the computer vision domain that have received limited attention in the existing literature. In addition, our work is centered around the utilization of automated annotation approaches, specifically ARESELP~\cite{xiong2017new} and a customized version of ARESELP (e.g., MARESELP). These techniques are then contrasted with the manual labeling method conducted by a domain expert.

% \section{Related Work}
The evaluation and validation of layer-tracking performance in ice sheets have been subjects of significant research interest. A common approach to assessing the performance of layer-tracking algorithms is the use of synthetic age-depth profiles~\cite{macgregor2015radiostratigraphy}, which involves generating a synthetic age-depth relationship using a one-dimensional Nye model. This model is used to intersect picked isochrones with a known age-depth profile, thereby assigning an age to each pick and propagating this age-depth relationship across the ice sheet.  The concept of isochrone connectivity is proposed by~\cite{karlsson2012continuity}. The metric assesses the degree of continuity and connectivity between detected ice layers, providing a quantitative measure of the uncertainty inherent in the tracking process. In addition, the metric is invaluable in highlighting areas where the interpretation consists of a high number of disconnected isochrones, which may indicate sensitivity to low amplitude signal anomalies.    

% \begin{table*}[ht!]
% \caption{Examples of binary mask arrays and their corresponding radargrams used in this study.}
%     \label{tab1}
%     \centering
%     \begin{tabular}{cccc}
%     \hline
%     Raw radargram&Manual by expert&ARESELP&MARESELP\\
%     \hline\hline
%     \multicolumn{4}{c}{20120508\_06\_016}\\ 
%     \begin{minipage}[c]{0.23\textwidth}\includegraphics[width=1\textwidth]{figures/20120508_06_016_raw.png}\end{minipage}&
%     \begin{minipage}[c]{0.23\textwidth}\includegraphics[width=1\textwidth]{figures/20120508_06_016_expert.png}\end{minipage}&
%     \begin{minipage}[c]{0.23\textwidth}\includegraphics[width=1\textwidth]{figures/20120508_06_016_areselp.png}\end{minipage}&
%     \begin{minipage}[c]{0.23\textwidth}\includegraphics[width=1\textwidth]{figures/20120508_06_016_mareselp.png}\end{minipage}\\
%     &&\\
%     \hline
%     \multicolumn{4}{c}{20120507\_07\_014}\\ 
%     \begin{minipage}[c]{0.23\textwidth}\includegraphics[width=1\textwidth]{figures/20120507_07_014_raw.png}\end{minipage}&
%     \begin{minipage}[c]{0.23\textwidth}\includegraphics[width=1\textwidth]{figures/20120507_07_014_expert.png}\end{minipage}&
%     \begin{minipage}[c]{0.23\textwidth}\includegraphics[width=1\textwidth]{figures/20120507_07_014_areselp.png}\end{minipage}&
%     \begin{minipage}[c]{0.23\textwidth}\includegraphics[width=1\textwidth]{figures/20120507_07_014_mareselp.png}\end{minipage}\\
%     &&\\
%     \hline
%     \end{tabular}    
% \end{table*}

\section{Material and Method}
\subsection{Radargram and Annotation Products}
In this study, we harness the extensive data resources provided by the Centre for Remote Sensing of Ice Sheets (CReSIS), accessing a publicly available repository, to develop our annotation products. Our focus is concentrated on a selected set of 100 radargrams, all sourced from various locations across North Greenland. 
These radargrams, carefully chosen for their diverse characteristics and representativeness, include notable sequences such as 20120330\_03\_019-028, 20120404\_01\_004-011, 20120507\_07\_003-015, 20120508\_04\\\_002-018, 20120508\_07\_001-012, 20120508\_07\_015-023, 20120511\_01\_041-052, 20120511\_01\_061-067, and 20120516\\\_01\_080-091. Utilizing these specific radargrams, we embark on a detailed process of generating annotations for the ice sheets. 
% These annotations are then meticulously transformed into binary mask arrays. This transformation is a critical step, as it facilitates a more structured and quantitative approach to studying the intricate features and variations within the ice sheets.
% Table~\ref{tab1} depicts the examples of binary mask arrays and their corresponding radargrams generated from three distinct approaches.

\subsection{Layer-Tracking Performance Metrics}
\subsubsection{Isochrones connectivity~\cite{karlsson2012continuity}}
This method meticulously evaluates both the connectivity and continuity of the layers that have been identified. It achieves this by quantifying three key aspects: the total number of picked layers ($\#TL$), the number of layers that are continuous and uninterrupted ($\#CL$), and the number of layers that exhibit discontinuities or breaks ($\#DL$). An effective layer annotator is characterized by its ability to maximize connectivity while minimizing discontinuity, thereby ensuring a more accurate and cohesive representation of the layers.

\begin{algorithm}
\caption{Dip Estimation and Comparison}
\label{dip}
\textbf{Require:} $mask^{a}$, $mask^{gt}$, window\_size\\
- Initialize $dip\_mask^{a}$ and $dip\_mask^{gt}$ as zero arrays with dimensions of $mask^{a}$ and $mask^{gt}$, respectively.\\
\textbf{for} each point in $mask^{a}$ and $mask^{gt}$\\
\quad- Select a window\_size.\\
\quad- Compute transitions in the window.\\
\quad- Calculate $y$ and $x$ differences of transitions.\\
\quad- Compute angles using $arctan2$ of $y$ and $x$ differences.\\
\quad- Calculate average dip as mean of angles.\\
\quad- Assign average dip to the corresponding point in dip\_results.\\
\textbf{end for}\\
- Calculate Pearson correlation coefficient between $dip\_mask^{a}$ and $dip\_mask^{gt}$.\\
\textbf{Return} Correlation coefficient $\rho$.
\end{algorithm}

\subsubsection{Vision-based Metrics}
\begin{itemize}
\item Pixel accuracy\\
It quantitatively evaluates the accuracy of two binary masks by comparing them pixel by pixel. The calculation is the sum of all matching pixels divided by the total number of pixels in one of the masks. It is formally specified as $Acc. = \frac{\sum_{i}^{N}(mask^{a}_{i}==mask^{gt}_{i})}{N}$, where $mask^{a}_{i}$ and $mask^{gt}_{i}$ correspond to the values of the $i^{th}$ pixel in the respective masks (i.e., ARESELP mask and ground truth mask), $N$ is the total number pixels in the mask, and $\sum_{i}^{N}(mask^{a}_{i}==mask^{gt}_{i})$ represents the sum of pixels where the two masks have identical values (either both pixels are 1 or both are 0).

\item Pearson's correlation of dip estimation\\
It measures the similarity between the dip datasets obtained from two binary masks. The dip of each mask is estimated by calculating the dip angle for each pixel within a specified window size, and averaging the angles derived from transitions in the binary data~\cite{panton2014automated}. Algorithm~\ref{dip} shows a pseudocode of the metric calculation.

\item Structural similarity index (SSIM)~\cite{wang2004image}\\
It assesses the similarity between two binary masks. It considers changes in texture, providing a more perceptually relevant assessment of layer annotation similarity compared to simpler metrics like mean squared error. Algorithm~\ref{ssim} shows a pseudocode for calculating SSIM.

\begin{algorithm}
\caption{Structural Similarity Index Calculation}
\label{ssim}
\textbf{Require:} $mask^{a}$, $mask^{gt}$\\
- Define parameters: window size, constants ($C_{1}$ and $C_{2}$) for stabilizing division with weak denominators.\\
- Initialize the SSIM map as a zero array with dimensions based on window size.\\
\textbf{for} each overlapping window in $mask^{a}$ and $mask^{gt}$.\\
\quad- Extract corresponding windows from $mask^{a}$ and $mask^{gt}$. \\
\quad- Calculate mean, variance, and covariance for these windows.\\
\quad- Calculate $y$ and $x$ differences of transitions.\\
\quad- Compute angles using $arctan2$ of $y$ and $x$ differences.\\
\quad- Compute SSIM for the current window:\\
       $\text{SSIM(window)} = \frac{(2 \mu_{x} \mu_{y} + C_1)(2 \sigma_{xy} + C_2)}{(\mu_{x}^2 + \mu_{y}^2 + C_1)(\sigma_{x}^2 + \sigma_{y}^2 + C_2)}$\\
\quad- Update SSIM map with computed SSIM value for the current window.\\
\textbf{end for}\\
- Calculate the mean SSIM over the entire image for the final SSIM value.\\
\textbf{Return} Final SSIM value.
\end{algorithm}

\item Recall IoU ($IoU_{r}$)\\
The Recall Intersection over Union (IoU) metric is a method used to evaluate the accuracy of binary masks, specifically focusing on the overlap between a predicted mask ($mask^{a}$) and a ground truth mask ($mask^{gt}$). It calculates the ratio of the overlapping area (where both masks agree on positive pixels) to the total area covered by the ground truth mask, providing a measure of recall or how well the predicted mask captures the relevant areas of the ground truth. Algorithm~\ref{riou} shows a pseudocode implementation of this metric.

\begin{algorithm}
\caption{Recall Intersection over Union Calculation}
\label{riou}
\textbf{Require:} $mask^{a}$, $mask^{gt}$\\
- Compute the overlap as the sum of element-wise logical AND between $mask^{a}$ and $mask^{gt}$.\\
- Compute the total number of positive pixels in $mask^{gt}$.\\
- Calculate Recall IoU as the ratio of overlap to the total layers in $mask^{gt}$.\\
$\text{Recall IoU} = \frac{\text{Overlap}}{\text{Total Layers in $mask^{gt}$}}$.\\
\textbf{Return} Recall IoU.
\end{algorithm}

\item Layer-by-layer Recall IoU ($IoU_{r}^{l}$)\\
It calculates the average recall for layer-by-layer comparison in binary masks. It first computes the IoU for each pair of layers between two masks. Then, it selects layer pairs with IoU scores above the average and calculates the recall for these pairs, which measures the proportion of actual positive samples (i.e., true layer matches) that are correctly identified. The average recall across these selected layers provides a metric for the overall accuracy of the layer identification in the masks. Algorithm~\ref{layeriou} shows the outline for calculating layer-by-layer recall IoU metric.

\begin{algorithm}
\caption{Layer-by-Layer Recall IoU Calculation}
\label{layeriou}
\textbf{Require:} iou\_scores, $mask^{a}$, $mask^{gt}$\\
- Initialize an empty list for recalls.\\
- Compute average IoU from iou\_scores.\\
- Select layer pairs from iou\_scores with IoU greater than or equal to the average IoU.\\
\textbf{for} each selected layer pair ($i$,$j$) in selected\_pairs\\
\quad- Extract corresponding layers from  $mask^{a}$ and $mask^{gt}$\\
\quad- Compute recall for the layer pair\\
\quad- Append recall to recalls list\\
\textbf{end for}\\
- Compute average recall from the recalls list\\
\textbf{Return} average recall
\end{algorithm}    
\end{itemize}

\begin{table}[ht!]
\caption{The mean quantitative score ($\mypm$ standard deviation) of all the layer annotation techniques, such as manual approach by the expert, ARESELP, and MARESELP in terms of isochrones connectivity.}
    \label{meanperf1}
    \centering
    \resizebox{0.49\textwidth}{!}{
    \begin{tabular}{p{5em}lll}
    \hline
    Method&$\#CL\uparrow$&$\#DL\downarrow$&$\#TL\uparrow$\\
    \hline\hline
    Manual&7.63$\mypm$6.75&144.03$\mypm$58.75&151.66$\mypm$59.99\\
    ARESELP~\cite{xiong2017new}&15.16$\mypm$8.48&39.11$\mypm$12.74&54.27$\mypm$17.02\\
    MARESELP&21.89$\mypm$9.55&56.03$\mypm$20.60&77.92$\mypm$25.57\\
    \hline
    \end{tabular}
    }
\end{table}

\begin{table*}[ht!]
\caption{The average quantitative score ($\mypm$ standard deviation) derived from all pairwise comparisons between the expert labels (ground truth) and the automatic annotation techniques, specifically ARESELP and MARESELP, using vision-based metrics.}
    \label{meanperf2}
    \centering
    \begin{tabular}{llllll}
    \hline
    Method&$\rho\uparrow$&$SSIM\uparrow$&$Acc.\uparrow$&$IoU_{r}\uparrow$&$IoU^{l}_{r}\uparrow$\\
    \hline\hline    ARESELP~\cite{xiong2017new}&0.527$\mypm$0.190&0.822$\mypm$0.059&0.973$\mypm$0.010&0.633$\mypm$0.185&0.313$\mypm$0.125\\
    MARESELP&0.547$\mypm$0.184&0.827$\mypm$0.056&0.974$\mypm$0.009&0.692$\mypm$0.154&0.325$\mypm$0.120\\
    \hline
    \end{tabular}
\end{table*}

\section{Result and Discussion}
% In this section, we delve into the comparative results of our study. 
The implementation of all metrics is readily accessible online to ensure reproducibility and facilitate further research. Table~\ref{meanperf1} provides a comprehensive comparison of different layer annotation techniques - manual annotation by experts, ARESELP, and MARESELP - focusing on their performance in terms of isochrone connectivity. The metrics used for this comparison include the number of continuous layers ($\#CL$), the number of broken layers ($\#DL$), and the total number of layers ($\#TL$), where a higher number of continuous layers and a higher total number of layers are desirable, while a lower number of broken layers is preferred. The manual approach, traditionally considered the gold standard due to its reliance on expert interpretation, shows a moderate number of continuous layers but a significantly high number of broken layers, resulting in a total layer count ($\#TL$) of 151.6$\mypm$59.99. This indicates the inherent challenges in manual annotation, where maintaining continuity across layers can be difficult, leading to a higher incidence of broken layers. In contrast, the ARESELP method shows a marked improvement in the number of continuous layers, more than double that of the manual approach. Notably, it also exhibits a substantial reduction in the number of broken layers, suggesting that this automated technique is more effective in maintaining layer continuity. The total number of layers identified by ARESELP is lower than that identified by the manual method, which may indicate a more selective layer identification process. MARESELP further advances these improvements, registering the highest number of continuous layers and a moderate number of broken layers, resulting in a total layer count ($\#TL$) of 77.92$\mypm$ 25.57. This suggests that MARESELP not only excels in identifying continuous layers but also strikes a balance in total layer detection, possibly offering a more nuanced and accurate representation of isochrones compared to the other methods. In summary, while the manual approach provides a substantial number of total layers, its high number of broken layers highlights the challenges of manual interpretation. ARESELP and MARESELP, on the other hand, demonstrate their strengths in automated layer annotation, particularly in maintaining layer continuity, as evidenced by their higher $\#CL$ and lower $\#DL$ scores. This evolution from manual to automated techniques underscores the potential of automatic approaches to enhance the accuracy and efficiency of ice sheet annotation.

In addition, Table~\ref{meanperf2} compares two automated annotation techniques, ARESELP and MARESELP, based on their performance in several vision-based metrics compared to expert labels (ground truth). These metrics include the Pearson correlation coefficient of the dip estimation ($\rho$), Structural Similarity Index ($SSIM$), Pixel Accuracy ($Acc.$), and Recall Intersection over Union for both global ($IoU_{r}$) and layer-by-layer ($IoU_{r}^{l}$) comparisons. The results demonstrate that while both ARESELP and MARESELP exhibit high accuracy and a strong structural resemblance to expert annotations, MARESELP shows a marginally better performance in aligning with expert labels, particularly in terms of IoU metrics. This indicates that MARESELP may offer a more refined and precise annotation than ARESELP, especially in complex layer-by-layer assessments. The superiority of MARESELP in terms of IoU metrics is particularly noteworthy, as it reflects a greater efficacy in capturing both the general and detailed aspects of ice sheet layers as annotated by experts However, annotation products from automatic approaches need to be further assessed, particularly when dealing with "hallucinated layers"- annotations that do not exist in the underlying radar imagery. 

\section{Conclusion}
This study presents various quantitative metrics to validate the performance of ice sheet annotation techniques, specifically manual, ARESELP, and MARESELP. Our analysis reveals that while manual annotation provides valuable expert insight, it struggles with layer continuity, a challenge effectively mitigated by automated methods. ARESELP shows marked improvement in maintaining layer continuity, and MARESELP further excels by achieving the highest number of continuous layers and a balanced total layer count. With respect to vision-based metrics, MARESELP marginally outperforms ARESELP, especially in IoU measures, indicating its closer alignment with expert annotations. These findings highlight the potential of automated annotation techniques in enhancing the accuracy and efficiency of ice sheet analysis, which is crucial for understanding ice dynamics and their impact on sea level changes in the context of climate research. 

\section{Acknowledgment}
This work is funded by the National Science Foundation (Institute for Harnessing Data and Model Revolution in the Polar Regions (iHARP) Award \#2118285).

\bibliographystyle{IEEEbib}
\bibliography{refs}

\end{document}